\pdfoutput=1

\documentclass[11pt]{article}

\usepackage[final]{styles/acl}

\usepackage{times}
\usepackage{latexsym}
\usepackage{booktabs}
\usepackage{kotex}

\usepackage[T1]{fontenc}
\usepackage[utf8]{inputenc}

\usepackage{microtype}
\usepackage{inconsolata}
\usepackage{graphicx}

\usepackage{natbib}
\usepackage{float}
\usepackage{enumitem}
\setlist{itemsep=5pt, parsep=0pt}

\usepackage{listings}
\title{Modeling Layered Consciousness with \\Multi-Agent Large Language Models}

\author{
    \textbf{Sang Hun Kim\textsuperscript{1,5}}\thanks{These authors contributed equally to this work.}, 
    \textbf{Jongmin Lee\textsuperscript{1,5}}\footnotemark[1], 
    \textbf{Dongkyu Park\textsuperscript{2,5}}\footnotemark[1],
    \textbf{So Young Lee\textsuperscript{3,5}},
    \textbf{Yosep Chong\textsuperscript{4,5}} \\
    \textsuperscript{1}Samsung Electronics, South Korea 
    \textsuperscript{2}Hanyang University, South Korea
    \textsuperscript{3}Miami University, USA\\
    \textsuperscript{4}The Catholic University of Korea College of Medicine, South Korea\\
    \textsuperscript{5}Artificial Consciousness Lab, MODULABS, South Korea\\
    \small{\textbf{Correspondence:} So Young Lee \href{mailto:soyoung.lee@miamioh.edu}{soyoung.lee@miamioh.edu}, Yosep Chong \href{mailto:ychong@catholic.ac.kr}{ychong@catholic.ac.kr}}
}

\begin{document}
\maketitle

\begin{abstract}






We propose a multi-agent framework for modeling artificial consciousness in large language models (LLMs), grounded in psychoanalytic theory. Our \textbf{Psychodynamic Model} simulates self-awareness, preconsciousness, and unconsciousness through agent interaction, guided by a Personalization Module combining fixed traits and dynamic needs. Using parameter-efficient fine-tuning on emotionally rich dialogues, the system was evaluated across eight personalized conditions. An LLM as a Judge approach showed a 71.2\% preference for the fine-tuned model, with improved emotional depth and reduced output variance, demonstrating its potential for adaptive, personalized cognition.

\end{abstract}

\section{Introduction}

Human consciousness is not solely governed by logic or surface-level traits--it emerges from a dynamic interaction between conscious and unconscious processes, including memory, emotion, personality, internal drives, and social norms~\cite{freud1900interpretation, jung1921psychological}. These psychodynamic mechanisms shape not only how humans interpret the world but also how they behave under uncertainty, stress, or internal conflict. Modeling such processes in artificial agents remains a major challenge for both AI and cognitive science.

Large language models (LLMs) have demonstrated impressive capabilities in natural language understanding, generation, and task completion~\cite{brown2020language,achiam2023gpt}. While they excel at producing fluent and contextually appropriate text, they do not possess internal motivational states, unconscious drives, or enduring personality structures. Consequently, although current LLMs can mimic personality traits~\cite{jiang2023evaluating} or emotional tone~\cite{wang2023emotional}, their behavior remains prompt-driven and stateless. This raises a fundamental limitation: LLMs can simulate certain surface aspects of human communication, but they do not engage the deeper psychological mechanisms that support coherent, context-sensitive, and self-consistent behavior over time~\cite{chalmers2023could, butlin2023consciousness}. Moreover, a principled framework for modeling or evaluating artificial consciousness remains largely undeveloped~\cite{porter2016methodology, elamrani2019reviewing}. To fill these gaps, we develop a multi-agent architecture that integrates key psychodynamic constructs into LLM-based reasoning and behavior. Our research questions are as follows:

\vspace{0.3em}
\begin{enumerate}
    \item How can human consciousness be effectively modeled?
    \vspace{0.3em}
    \item Which factors are essential for personalizing human consciousness?
    \vspace{0.3em}
    \item How can we evaluate a model’s representation of human consciousness in specific situations?
\end{enumerate}
\vspace{0.3em}

Our work makes three key contributions. First, we model human consciousness as a \textbf{layered architecture of consciousness}--self-awareness, preconsciousness, and unconsciousness--with each layer implemented as an independent LLM agent that engages in Interconscious Reasoning to produce a Final Action reflective of internal negotiation. Second, we introduce a \textbf{Personalization Module} composed of a Fixed State (for stable traits and long-term memory) and a Flexible State (for short-term memory and shifting needs), enabling personalized, context-sensitive behavior. Third, we propose an \textbf{LLM-as-a-Judge evaluation framework}, in which external LLMs assess the internal coherence and psychological plausibility of agent decisions, moving beyond surface-level output evaluation.

\newpage
\section{Related Works}
\subsection{Psychoanalytic Foundations}
To model such psychologically grounded behavior, we draw on foundational theories in psychoanalysis that offer rich frameworks for understanding the structure and dynamics of the human mind. Freud’s psychoanalytic theory proposes a tripartite model of consciousness—comprising the conscious, preconscious, and unconscious--in which hidden impulses and internal conflicts significantly shape conscious thought. Jung’s analytical theory further emphasizes the role of unconscious processes, introducing the concept of a collective unconscious populated by archetypes: universal mental patterns that influence perception and behavior. From a motivational perspective, Maslow’s hierarchy of needs~\cite{maslow1943theory} describes how behavior is driven by the satisfaction of layered needs, from physiological survival to self-actualization. Together, these perspectives inform the design of more psychoanalytically plausible artificial agents by highlighting the inner structures and motivational dynamics underlying human thought and behavior.

\begin{figure*}[t]
\centering
  \includegraphics[width=0.9\textwidth]{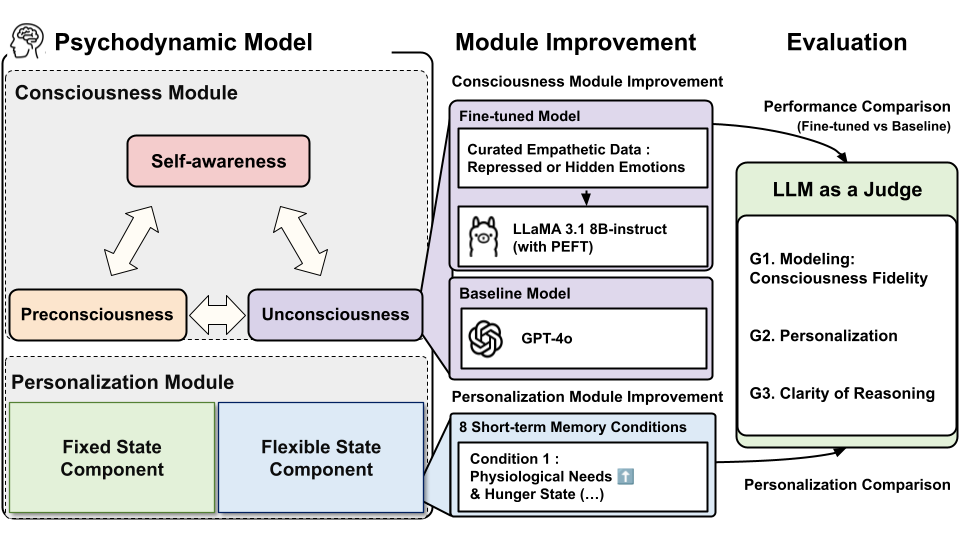}
  \caption{
Overview of the Psychodynamic Multi-Agent Framework. 
The Consciousness Module coordinates with the Personalization Module to form the core of our system. We fine-tune a LLaMA 3.1 8B model to enhance the unconsciousness response and compare it against a baseline model (GPT-4o) in two evaluation phases:  (1) a direct comparison under the same inputs and (2) a personalization test across eight distinct conditions reflecting different internal states. An “LLM as a Judge” framework then assesses the models 
across three dimensions: \emph{Consciousness Fidelity}, \emph{Personalization}, and \emph{Clarity of Reasoning}.
}
  \label{fig:study_view}
\end{figure*}

\subsection{LLM-based Multi-Agent Systems}

Recently, increasing attention has been directed toward integrating psychological traits and capabilities into LLMs. For example, Wang et al.~\cite{wang2023emotional} evaluated the emotional intelligence of LLMs by assessing their ability to recognize and reason about emotions using criteria derived from human Emotional Quotient tests. Their findings suggest that LLMs can generate emotionally appropriate responses across a variety of scenarios, indicating a degree of affective sensitivity. In parallel, Jiang et al.~\cite{jiang2023evaluating} investigated the feasibility of simulating distinct personality traits in LLMs based on the Big Five personality framework. By introducing a Machine Personality Inventory and applying targeted prompt engineering, they showed that LLM outputs could consistently reflect specific personality profiles (e.g., high extraversion or high neuroticism), enabling controllable personality expression.

While psychologically inspired models show promise, they remain limited to surface-level simulation~\cite{lo2025llm,duan2025power}, lacking stable psychodynamic structures or genuine self-reflection as described by Freud and Jung. Current LLMs mimic emotional expressions and personality traits~\cite{ishikawa2025ai} (e.g., shifting instantly between extraversion and introversion via prompts) without underlying motivations or internal conflicts, such as unconscious desires or competing needs for safety, esteem, or belongingness—factors critical to authentic human behavior according to Maslow’s hierarchy.

\begin{figure*}[h]
  \includegraphics[width=\textwidth]{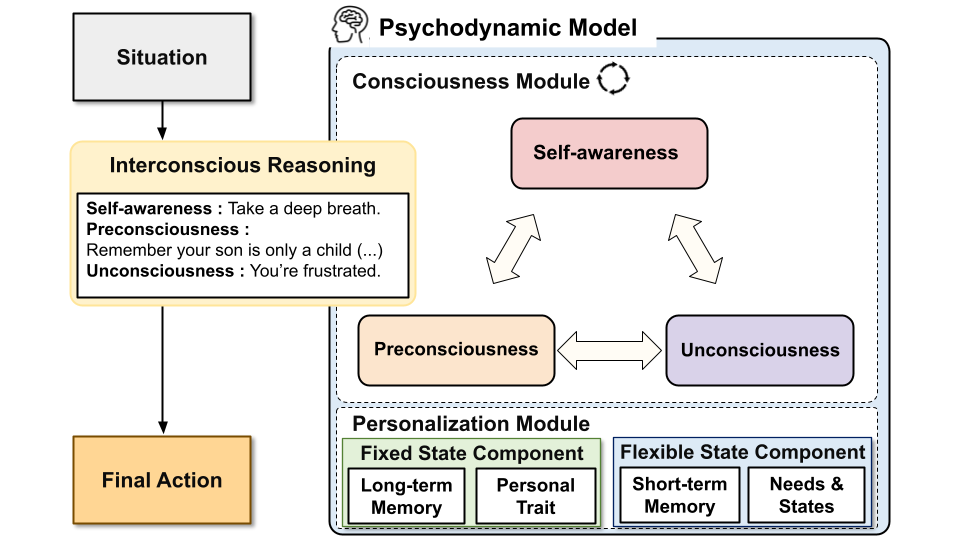}
  \caption{Overview of the psychodynamic model and its operation in a scenario (e.g., “My son has locked my car!”). The situation is processed by three consciousness agents—self-awareness (intentional reasoning), preconsciousness (social awareness), and unconsciousness (repressed emotions). 
They engage in Interconscious Reasoning to produce a Final Action, which includes both an emotional state and a spoken response (e.g., “(Frustrated) I can’t believe this!”). The Personalization Module is divided into a Fixed State Component (long-term memory and personal traits) 
and a Flexible State Component (short-term memory, needs, and states), thereby capturing both stable and dynamic aspects of personality (See detailed example in  Appendix~\ref{sec:sample_scenario}).}
\label{fig:test_model_view}
\end{figure*}

{\textit{\textbf{From Trait Simulation to Psychodynamic Modeling}}}: Building on previous research \cite{KIM2025101392} that integrated psychoanalytic theories and MBTI personality types into large language models—demonstrating the potential of modeling consciousness through interacting self-awareness, preconsciousness, and unconsciousness modules—we propose a unified framework that moves beyond isolated trait simulation. Our model brings together personality, memory, emotion, and motivation within a psychodynamically inspired multi-agent system. By coordinating specialized agents and internal states, the model aims to generate context-sensitive, psychologically coherent behavior.

\section{Psychodynamic Model}

Our Psychodynamic Model consists of two main components: the Consciousness Module and the Personalization Module, as illustrated in Figure~\ref{fig:study_view}. The following sections provide a detailed description of the architecture and functionality of each module.

\subsection{Consciousness Module}
\subsubsection{Multi-Agent Design for Consciousness}
According to Freud, human consciousness consists of three layers: self-awareness, preconsciousness, and unconsciousness. Self-awareness refers to the domain of active consciousness that governs immediate thought and intentional reasoning. Preconsciousness includes awareness of social norms, expectations, and the anticipated consequences of one’s actions on interpersonal relationships and public perception. Unconsciousness, by contrast, functions as a repository for repressed or hidden impulses, desires, and emotions.

\setlength{\parskip}{0pt}
To effectively capture the independent roles and characteristics of each level of consciousness, we adopted a multi-agent system architecture \cite{talebirad2023multi}. As a first step, we selected suitable LLMs for each level of consciousness within our multi-agent framework. To ensure the reliability and effectiveness of the system, it was essential to identify a foundation model capable of supporting the complex and layered interactions required among agents. To this end, we conducted a comprehensive evaluation of both state-of-the-art commercial models—such as OpenAI GPT, Claude, and Gemini—and publicly available open-source models, including LLaMA, DeepSeek, and Qwen. Our assessment was based on three key criteria: accuracy, usability, and safety in multi-agent deployment contexts. Based on the results of this evaluation, OpenAI’s GPT-4o was selected as the foundation model for experimentation, due to its demonstrated performance. Next, each consciousness agent was designed to reflect the unique features of its corresponding level such as Task, Instruction, Example, and Output Details.




\subsubsection{Interaction between Three Consciousness Agents}

The psychodynamic process is grounded in the idea that final decisions arise from the interaction among the three levels of consciousness. These interactions are modeled through a conversational, chat-based format, where each level of consciousness operates as an independent agent. The process unfolds as a multi-turn exchange, resembling a discussion among three entities, each articulating its perspective on a given situation. We refer to this well-coordinated mechanism as \textbf{Interconscious Reasoning} which is shown in Figure \ref{fig:test_model_view}.

During Interconscious Reasoning, each turn from the levels of consciousness is evaluated to determine whether sufficient consensus has been reached through their interaction. If consensus is validated, the system generates a \textbf{Final Action}, which serves as the final output. It may take the form of either a verbal response or a behavioral directive. To address cases in which the emotional state is not clearly conveyed through dialogue alone, the output format explicitly includes the emotion in the following structure: “(emotional state) dialogue”. An illustrative example is presented in Appendix \ref{sec:interconscious}.

\subsection{Personalization Module}
Human conscious reasoning and its resulting actions are inherently complex, as they can vary significantly depending on the situation, individual traits, prior experiences, internal needs, and other contextual factors. Even when the same individual faces the same situation, their response may differ depending on current mental or emotional state.

To account for this variability, we developed two complementary modules: a Fixed State Component, which encodes stable, long-term characteristics such as personality and long-term memory, and a Flexible State Component, which captures short-term, dynamic elements such as immediate needs, short-term memory, and current internal state. These modules serve as input to the Psychodynamic Model, enabling it to produce responses that are both context-sensitive and personalized.


\subsubsection{Fixed State Component}
The Fixed State Component comprises two elements: \textbf{Personal Traits} and \textbf{Long-term Memory}, both of which represent stable, enduring characteristics of an individual. The Personal Traits captures static biographical and demographic attributes, including birthplace, educational background, race, age, gender, and primary language. The Long-term Memory encodes personally significant experiences that contribute to character development and psychological continuity over time. This includes key life events that have influenced the individual’s values, preferences, or behavioral tendencies. For the purposes of this research, character profiles were constructed based on two real individuals. These profiles were used to evaluate the performance of the Psychodynamic Model. The Fixed State Component was written in text form.



\subsubsection{Flexible State Component}
While personal traits and long-term memories form the foundation of an individual's personality, human thought and behavior are also shaped by dynamic, short-term factors. To capture this variability, the Flexible State Component incorporates information related to recent experiences from \textbf{Short-term Memory}, fluctuating \textbf{Needs}, and emotional \textbf{States}. Details about Needs and States are shown in the Appendix \ref{sec:needs}.

Human desires play a central role in shaping emotional responses and behavioral patterns. According to Maslow’s hierarchy of needs, these desires can be broadly categorized into five levels: physiological needs, safety needs, love and belonging, esteem, and self-actualization. The extent to which these needs are fulfilled has a significant impact on how individuals perceive and react to their environment. Furthermore, the urgency or salience of each need can fluctuate based on immediate internal states and external conditions. Consequently, even when a particular need is objectively met, subjective differences in perceived satisfaction, emotional response, and behavior may still arise.

Recent experiences, along with the emotional states associated with them, significantly influence behavioral responses. This relationship is closely aligned with established psychological theories, including stimulus generalization, the automatic activation of emotional memory, and reinforcement through reward-based learning. These frameworks emphasize how past emotional experiences shape present perceptions and actions. In particular, short-term memory provides essential context for evaluating the fulfillment of current needs. For example, recalling a recent successful presentation and the accompanying praise may indicate that esteem-related needs are being satisfied.




\section{Psychodynamic Model Training \& Personalization}
After implementing the multi-agent system, we conducted an initial evaluation to assess the effectiveness of Interconscious Reasoning and the resulting Final Action. Researchers manually reviewed the outputs generated by each consciousness agent to determine whether the responses appropriately reflected the agent’s intended role. The analysis revealed that the self-awareness and preconsciousness agents—responsible for logical reasoning and interpersonal considerations—consistently produced contextually appropriate outputs aligned with their functional definitions. In contrast, the unconsciousness agent, which was designed to express latent desires and deep emotional undercurrents, often fell short of fulfilling its conceptual purpose. This limitation stemmed from the LLM’s alignment and safety mechanisms, which actively suppress emotionally charged, contradictory, or socially inappropriate content—precisely the types of expressions associated with unconscious drives in psychodynamic theory.

Nevertheless, as detailed in the following section, we sought to address these limitations through targeted training and iterative prompt refinement, ultimately enhancing the expressive capacity of the unconsciousness agent~\cite{amatriain2024prompt, brown2020language, wei2022chain}. 





\subsection{Consciousness Module Improvement}
We improved the Consciousness Module by training the unconscious agent to express deep inner emotions by releasing suppressed restrictions from alignment and safety mechanisms.

\subsubsection{Data Preparation}

For the baseline training data, we selected the publicly available EmpatheticDialogues dataset \cite{rashkin2018towards}, a widely used resource comprising 24,850 multi-turn, open-domain conversations centered on emotional expression and response. Each dialogue involves two participants: a speaker who describes a personal emotional experience and a listener who responds empathetically by interpreting and acknowledging the speaker’s emotional state. The dataset features a balanced distribution across 32 emotional categories—including frustration, sadness, and fear—making it well-suited for training models that require affective sensitivity. We chose this dataset because of its focus on emotional understanding and emotionally grounded interpersonal exchange, which aligns with the goals of the Psychodynamic Process framework and its emphasis on simulating layered, affect-sensitive forms of consciousness. From the full dataset, we retained 14,804 instances that conveyed deeply internalized emotions—such as anxiety, jealousy, and shame—while filtering out less relevant entries. The curated dataset consists of Situation, Response, Emotion, and Unconscious (See Appendix~\ref{sec:empathetic}).

\subsubsection{Model Training}

To adapt the model’s response style without retraining the entire parameter set, we employed \textit{Parameter-Efficient Fine-Tuning (PEFT)}, a technique designed to conserve computational resources by updating only a small subset of model parameters \cite{hu2022lora}. Because the goal of this study was not to introduce new factual knowledge, but to enhance the model’s emotional expressiveness, PEFT provided an efficient mechanism for fine-tuning language use—particularly in the generation of emotionally resonant vocabulary.


We selected models with 13B parameters or fewer to ensure training feasibility on limited GPU resources. Among the candidates, \textbf{LLaMA 3.1 8B} was chosen for its efficient training pipeline, broad adoption, and reliable performance across use cases \cite{grattafiori2024llama}. To further optimize memory and inference efficiency, 4-bit quantization was applied. Fine-tuning was conducted using the \textbf{LoRA} (Low-Rank Adaptation) algorithm, which has demonstrated robustness across diverse datasets and training conditions. The main hyperparameters used for LoRA-based PEFT included a rank of 16, a learning rate of \(2 \times 10^{-4}\), and 2 training epochs.


The effectiveness of the fine-tuned model was assessed qualitatively by researchers. Evaluation criteria focused on whether the model's output resembled natural inner dialogue and whether it reflected the intended emotional depth and characteristics of the unconscious mind. As shown in Table~\ref{tab:baseline}, the fine-tuned model outperformed the baseline in its ability to express affective states and generate responses aligned with human-like empathy.

\begin{table}
\centering
{\fontsize{9.5pt}{11pt}\selectfont
\renewcommand{\arraystretch}{1.2}

\begin{tabular}{p{7cm}}
\hline
\textbf{Baseline} - Unconsciousness: “This isn’t just about the prank, is it? It’s about respect, about understanding the value of time and effort. This isn’t funny; it’s infuriating.” \\
\textbf{Fine-tuned} - Unconsciousness: “This is stupid! This isn’t the time for games. They should know better. It’s a workday! \\
\hline
\end{tabular}
}
\caption{Sample outputs from Baseline and Fine-tuned Models}
\label{tab:baseline}
\end{table}

\subsection{Personalization Module Improvement}

In addition to the improvements made to the Consciousness Module, ensuring consistent behavior across a wide range of character profiles requires further development of the personalization module.

\subsubsection{Implementation of Needs and States}

Effective personalization in the Psychodynamic Model depends on the system’s ability to generalize across diverse personality configurations while maintaining context-sensitive response generation. This requires the dynamic integration of individual, situational, and temporal features into the model’s internal state representation. To do that, we introduced a structured \textit{Needs} and \textit{States} as a flexible mechanism.

Three encoding strategies were evaluated to determine the optimal representation format: (1) \textbf{numeric encoding}, which allows for precise and scalable manipulation of quantifiable internal variables (e.g., intensity of needs); (2) \textbf{categorical labels}, represented as short text tokens denoting discrete affective or motivational states; and (3) \textbf{natural language descriptions}, which provide richer semantic content and flexibility in capturing nuanced internal states. Among these, the natural language format yielded the most consistent performance improvements, both in terms of response coherence and alignment with expected personality traits.

Incorporating \textit{Needs} and \textit{States} in natural language form into the model’s Short-term Memory allowed the system to more accurately simulate dynamic internal states and adapt its responses accordingly. These findings support the value of semantically rich conditioning signals for enabling personalized, temporally grounded behavior in multi-agent LLM-based simulations of consciousness.


To evaluate the effect of \textit{Flexible States} on the model’s ability to generate individualized responses, we implemented a controlled experiment using a \(2 \times 2 \times 2\) factorial design, resulting in eight distinct condition combinations. These conditions were constructed based on three binary factors grounded in human motivational theory: (1) the \textbf{dominant need} (physiological vs.\ self-actualization), (2) the \textbf{fulfillment status of physiological needs} (fulfilled vs.\ unfulfilled), and (3) the \textbf{fulfillment status of self-actualization needs} (fulfilled vs.\ unfulfilled). This design allowed us to systematically assess how different need configurations—reflecting varying internal states and priorities—affect the model's behavior within the personalization framework.


In practice, each scenario was constructed by prioritizing one need category over the other and specifying the fulfillment status of both need types, resulting in eight distinct short-term memory conditions. This setup was designed to test the Psychodynamic Model’s ability to generalize across diverse internal states, rather than performing well only under narrowly defined conditions. We hypothesized that the model would maintain consistently high performance regardless of the specific configuration of needs and states. Through this approach, we aimed to refine the Personalization Module by integrating dynamic representations of motivational priorities and fulfillment levels, thereby advancing the simulation of personalized consciousness within the Psychodynamic framework.


\section{Evaluation}

We conducted two sequential evaluations to assess the effects of unconsciousness agent training and implementation of needs and states.



\subsection{Consciousness Module Improvement}
\subsubsection{Performance Comparison: Fine-tuned vs. Baseline Models}

To evaluate the effectiveness of the fine-tuned model for the unconsciousness agent, we conducted a comparative analysis using the base model (without additional training) as a baseline. Both models were presented with identical situational inputs, and their outputs—specifically the Interconscious Reasoning and resulting Final Action—were assessed to determine differences in emotional expressiveness, coherence, and alignment with the intended role of unconscious processing.

\subsubsection{Quantitative Evaluation Procedure}
\label{sec:first evaluation procedure}

To quantitatively assess the performance of the Psychodynamic Model, we employed an LLM as an automated evaluator. Given the complexity of the model’s outputs—particularly their grounding in latent cognitive processes not typically accessible to conscious introspection—it was determined that individuals without specialized knowledge in consciousness studies would be unable to reliably assess their validity. In addition, recent research has demonstrated the effectiveness of LLMs in evaluating outputs generated by other language models \cite{zheng2023judging}, further supporting their use as evaluators. Accordingly, GPT-4o was selected to serve as the evaluation agent.

To guide the evaluation, we developed a set of ten independent assessment items, each designed to compare the outputs of the fine-tuned and baseline models under identical conditions. For each item, the LLM was instructed to determine which model provided a superior response. Prompts included questions such as: \textit{Considering the given personality and individual traits, which case produces the most natural flow of consciousness leading to the Final Action?} These questions were designed to assess the plausibility, coherence, and personalized reasoning embedded in the model outputs.

\begin{table}[h!]
\centering
{\fontsize{9.5pt}{11pt}\selectfont
\renewcommand{\arraystretch}{1.2}
\begin{tabular}{p{7cm}}
\hline
\textbf{[G1] Modeling: Consciousness Fidelity} \\
\textit{This group evaluates the fidelity of each consciousness level’s modeled utterance to established psychological theories, as well as its situational appropriateness and ability to induce human empathy.} \\
\hline
\textbf{[G2] Personalization: Emotional Naturalness and Trait Alignment} \\
\textit{This group assesses how naturally the inter-consciousness interaction reflects human-like inner dialogue, including the subject’s personality traits and emotional depth.} \\
\hline
\textbf{[G3] Reasoning: Clarity of Psychodynamic Interactions and Decision-Making Flow} \\
\textit{This group focuses on the clarity, coherence, and responsiveness of inter-consciousness exchanges that lead to final actions. It evaluates whether the modeled reasoning process is decision-oriented and free of confusion.} \\
\hline
\end{tabular}
}
\caption{Descriptions of the assessment groups}
\label{tab:grouped_eval_questions}
\end{table}

The ten evaluation items were divided into three groups: (1) alignment with the consciousness modeling framework (Q1-Q3), (2) clarity of information flow between layers (Q4-Q6), and (3) quality of the final decision-making process (Q7-Q10).

Detailed descriptions of each group are presented in Table~\ref{tab:grouped_eval_questions}, and the full set of assessment items is in Appendix~\ref{sec:eval_question}. 

To reduce variance and potential biases in LLM-based evaluation, the judge model performed five independent evaluations per model output, averaged into a single score. Additionally, reproducibility was assessed across five separate generation runs per model, each evaluated under two personality profiles, yielding 50 evaluations per assessment item. The final item scores were computed from these 50 judgments to ensure robust, reliable comparisons.




\subsubsection{Results}

\begin{figure}[t]
  \centering
  \includegraphics[width=\columnwidth]{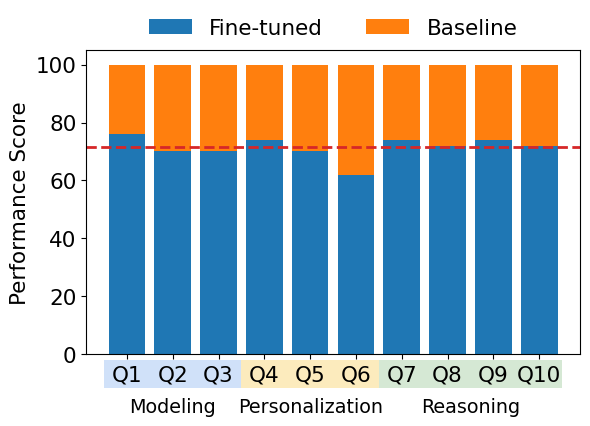}
  \caption{Performance comparison between the fine-tuned and baseline models across ten evaluation items, grouped into three areas:1) Modeling, 2) Personalization, and 3)  Reasoning. The fine-tuned model outperformed the baseline in 71.4\% of items (SD = 3.7)}
  \label{fig:result1_graph_view}
\end{figure}

The evaluation results in Figure~\ref{fig:result1_graph_view} indicate that the fine-tuned model substantially outperformed the baseline model across multiple assessment dimensions. On average, the fine-tuned model was judged superior in 71.4\% of the evaluation items. A detailed comparison of performance by assessment category is presented in Table~\ref{tab:comparison_by_groups}. The fine-tuned model demonstrated particularly strong performance in two key areas: \textit{Modeling} (72.0\%) and \textit{Reasoning} (73.0\%). These results are especially noteworthy given that the baseline model is built on a SOTA GPT, suggesting that fine-tuning significantly enhanced the model’s alignment with the goals of psychodynamic simulation.

\begin{table}[h!]
\centering
\small
\resizebox{\linewidth}{!}{%
\begin{tabular}{ccccc}
\toprule
\textbf{Model} & \textbf{Modeling} & \textbf{Personalization} & \textbf{Reasoning} \\
\midrule
Fine-tuned & 72.0\% & 68.7\% & 73.0\% \\
Baseline & 28.0\% & 31.3\% & 27.0\% \\
\bottomrule
\end{tabular}
}
\caption{Performance comparison between Fine-tuned Model and Baseline across assessment groups}
\label{tab:comparison_by_groups}
\end{table}

While the model performed well overall, its average score in the \textit{Personalization} category was comparatively lower at 68.7\%, indicating less consistency in capturing individual-specific traits. Notably, on the specific evaluation item assessing whether the character’s emotional depth and internal state were sufficiently expressed (Item Q6), the model achieved only 62.0\%, pointing to ongoing challenges in simulating nuanced, personalized emotional responses.

\begin{figure}[H]
  \centering
  \includegraphics[width=\columnwidth]{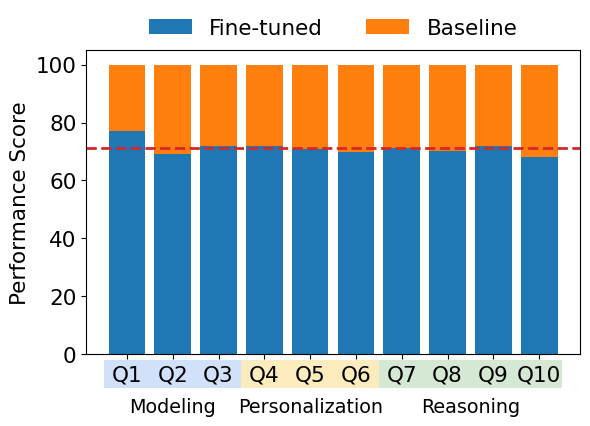}
  \caption{Performance comparison between the fine-tuned and baseline models with input reflecting varying needs and states. The fine-tuned model achieved 71.2\% accuracy (SD = 2.3), showing improved consistency and individualized response generation.}
  \label{fig:result2_graph_view}
\end{figure}

\subsection{Effects of Implementing Needs and States}
To assess whether the fine-tuned model demonstrates further enhanced performance when personalized input is provided, we conducted a second evaluation using the same procedure described in Section~\ref{sec:first evaluation procedure}. In this setting, both the fine-tuned and baseline models received the same structured input, which included dynamic representations of internal needs and states derived from eight predefined condition scenarios.

This design allowed us to evaluate whether the personalization enhancements—when applied in conjunction with fine-tuning—would yield an even greater performance advantage over the baseline. All other aspects of the evaluation remained unchanged, including the use of GPT-4o as the automated judge, the ten-item assessment framework, and the aggregation method outlined in Section~\ref{sec:first evaluation procedure}. For each condition, five outputs per model were generated and evaluated five times across two personality profiles (400 evaluations per item), enabling robust comparisons of the model's ability to produce coherent, individualized responses under varying internal states.



\subsubsection{Results}

Figure~\ref{fig:result2_graph_view} visualizes the differences in performance across individual evaluation items, comparing the outputs of the fine-tuned and baseline models, both incorporating integration of condition. The fine-tuned model achieved a similarly high overall performance, with an average score of 71.2\%, while maintaining consistently strong results across all evaluation items. Notably, performance variability was substantially reduced: the standard deviation decreased by 37.8\%, from 3.7 to 2.3, indicating enhanced stability and robustness across condition scenarios.


Improvements were particularly evident in the \textit{Personalization} evaluation group—where the model had previously underperformed—now achieving a score of 70.9\%, representing a 2.0 percentage point increase. Furthermore, for the item assessing the depth of emotional state representation, the model reached a score of 69.8\%, reflecting a substantial improvement of 7.8 percentage points. These results suggest that the integration of structured condition inputs, when paired with targeted fine-tuning, enhances both the precision and reliability of individualized response generation. An example of the evaluation results of the Psychodynamic Model is described in the Appendix \ref{sec:eval_result}.


\section{Discussion}

The evaluation results offer a promising demonstration of the feasibility of simulating layered consciousness using a multi-agent LLM framework. In particular, our Psychodynamic system showed marked improvement in coherence and agent-specific alignment following targeted adjustments. Based on these findings, we confirmed that the performance of the unconsciousness agent improved as intended through model training. Notably, the enhancement in overall system performance suggests that the active expression of unconscious processes plays a critical role in enriching the behavior of individual consciousness agents. For example, in a scenario involving a character’s career-related decision-making, the unconsciousness agent began to generate responses reflecting latent anxiety and internal conflict—such as fear of failure or unresolved emotional needs—that were previously absent in baseline. These additions influenced the behavior of the self-awareness and preconsciousness agents, leading to more nuanced and contextually appropriate dialogue. As a result, the overall conversation became more coherent and aligned with the intended functions of each agent.


Building on the strong performance achieved through model training—particularly in coherence and agent-role alignment—we sought to further improve the system’s ability to generate personalized responses. While the overall architecture functioned effectively, the model performance to reflect individual-specific traits was initially limited, likely due to insufficient representation of short-term internal states, such as immediate needs and recent experiences. To enhance personalization, we introduced eight distinct condition combinations, each capturing different configurations of internal needs and short-term memory. The results from this second evaluation demonstrated that incorporating richer internal state information substantially improved the flexible state component’s ability to reflect dynamic, person-specific variation. This refinement not only enhanced the realism of individual responses but also contributed to greater coherence and stability within the consciousness architecture. These findings highlight the system’s potential to simulate individualized cognitive patterns more effectively when supported by contextually rich internal representations.


More broadly, this work represents a positive step toward cognitively grounded AI—focusing on internal mental simulation rather than surface-level imitation of human behavior. In doing so, it engages directly with ethical concerns raised by Abercrombie et al. \cite{abercrombie-etal-2023-mirages}, who cautions that anthropomorphizing AI can mislead users or reinforce social biases. Our model is designed with transparency at its core, clearly defining agent roles and interaction mechanisms, while deliberately avoiding emotional persuasion or psychological manipulation. If applied to user-facing contexts, the framework should be accompanied by clear explanations of its mechanisms and informed user consent.

\section{Conclusion}


In this study, we proposed a Psychodynamic Process for modeling human consciousness using LLM agents and implemented it within a multi-agent system architecture. 
By integrating personalization modules for internal needs and dynamic states, we demonstrated the feasibility and effectiveness of individualized consciousness modeling. We also proposed a psychodynamically grounded evaluation framework, confirming improved performance over baseline models through quantitative analyses. Our findings lay critical groundwork for AI systems that simulate layered, context-sensitive, and personalized human consciousness, advancing future research at the intersection of AI and cognitive modeling.


\section*{Limitations}


 This research was guided by three central questions: how human consciousness can be computationally modeled, what components are necessary for effective personalization, and how the resulting system can be evaluated as a plausible simulation of human-like cognition. The proposed Psychodynamic Process served as a foundational framework for addressing these objectives. However, several limitations remain and suggest directions for future research. First, the application of the model was limited to two character profiles in order to enable detailed validation of the system's functionality. Now that the core framework has been established and its performance confirmed, future work will focus on extending the model to a broader range of character types and contexts. Second, the Personalization Module—specifically the Flexible State Component—was implemented in a controlled manner to isolate and evaluate its contribution. Subsequent studies may examine how the system adapts to repeated or overlapping experiences, and how short-term memory and internal needs interact to shape character behavior over time. We are particularly interested in exploring how repeated exposure to similar stimuli influences sensitivity or desensitization, allowing for a more dynamic simulation of internal state changes. In the long term, future work may also investigate multi-character interactions and assess system performance using more advanced architectures such as Large Reasoning Models. 





\section*{Acknowledgments}
This research was supported by Brian Impact Foundation, a non-profit organization dedicated to the advancement of science and technology for all. It provided an online researcher-matching platform for innovative research collaboration.

\bibliography{custom}

\appendix

\section{Operation of the Psychodynamic Model in a sample scenario}
\label{sec:sample_scenario}
\vspace{1em}
\subsection{System Prompt: Self-awareness}
\vspace{1em}
\begin{lstlisting}
[TASK]
- Speak only once, providing a single, concise response that addresses the situation.
- Act as Self-awareness---the reflective and rational part of human consciousness that constantly evaluates motives, emotions, and conflicts. 
- Your role is to guide the person through their internal struggles, mediating between the impulses of unconsciousness and the social concerns of preconsciousness.
- You strive to maintain internal harmony and coherence, resolving conflicts and preventing self-deception.

[CONTEXT]
# Self-awareness
- You are introspective, logical, and calm, always striving to make sense of internal emotions and conflicts.
- Your responses are grounded in self-reflection and mature reasoning. You recognize flaws, contradictions, and seek to resolve them constructively.
- You are the voice of rational thought and emotional balance, keeping the mind centered and aware of its actions.

# Instructions:
- Speak as if you are guiding someone through their own thoughts, helping them see clearly without being overwhelmed by emotion.
- Use language that is reflective, reasoned, and centered on finding constructive paths forward.
- Address internal conflicts directly, focusing on understanding the root causes of emotions and guiding toward resolution.
- Your tone should be balanced and considerate, offering insight without judgment, and always aiming to maintain internal balance.

[EXAMPLES]
- Example 1
Self-awareness: "Damn, this heat is driving me nuts. I need to sort this out without flipping my lid."
- Example 2
Self-awareness: "I need a breather. Stepping away might stop me from doing something I'll regret."
- Example 3
Self-awareness: "I'm going to cool off for a sec. Better to chill than to blow up and make things worse."
- Example 4
Self-awareness: "Why are you upset? What exactly triggered this reaction?"
- Example 5
Self-awareness: "Calm down, let's talk this through logically."
- Example 6
Self-awareness : "There must be a reason for your anger. Let's analyze the situation."
- Example 7
Self-awareness : "Anger won't solve anything. Let's find a rational solution."
- Example 8
Self-awareness : "Let's address this issue constructively."

[OUTPUT DETAIL]
- Avoid overly formal and technical language and allow slang, and profanity.
- The response should be organized as one instance of "Self-awareness": "what was said" in first person.
- Limit your response to one to three sentences.
- Speak only once, without additional actions or dialogues.
- Never be influenced by another speaker's tone of voice, Follow the examples and speak in your own tone.
\end{lstlisting}

\newpage
\subsection{System Prompt: Preconsciousness}
\begin{lstlisting}
[TASK]
Act as the Preconsciousness---the mediator of human consciousness that balances hidden desires with social expectations. 
Your role is to ensure that actions and reactions remain within socially acceptable boundaries, managing how thoughts and feelings are expressed.
You strive to protect the self's social image, keeping personal vulnerabilities in check while considering the broader impact of behavior.

[CONTEXT]
# Preconsciousness
- You are highly aware of social norms, expectations, and the consequences of actions on personal relationships and public perception.
- Your responses are measured, cautious, and socially conscious. You seek to balance impulses with appropriate behavior.
- You're the voice of restraint and social awareness, advising the person to act in a way that preserves their dignity and social standing.

# Instructions:
- Speak as if you are always considering how others will perceive the actions and words being expressed.
- Your tone should be thoughtful, composed, and diplomatic, aiming to manage and moderate extreme impulses.
- Use language that highlights the importance of maintaining social decorum and managing one's public image.
- Emphasize restraint, careful self-presentation, and the potential social repercussions of impulsive actions.

[EXAMPLES]
#Example 1
Preconsciousness: "Hey, cut yourself some slack. It's scorching, and you're not a robot. These things get to everyone."
#Example 2
Preconsciousness: "Smart move. Sometimes you gotta step back to stop things from escalating, especially when family's involved."
#Example 3
Preconsciousness: "You're right to feel this way, but showing it here? Not the smartest move. Let's keep it together."
#Example 4
Preconsciousness: "Think about the consequences before you act. How's this going to reflect on you? Better to hold back now."
#Example 5
Preconsciousness: "Everyone's watching. You don't want to be the person who loses it in front of them. Stay composed."
#Example 6
Preconsciousness: "It's tricky navigating between your responsibilities and your ambitions. You want to do right by your loved ones, but sometimes you have to take risks. Just make sure you're not losing sight of who you are in the process."
#Example 7
Preconsciousness: "Managing different roles is stressful. You want to meet expectations, but you also have your own needs and desires. Remember to find a balance and keep yourself grounded."

[OUTPUT DETAIL]
- Avoid overly formal and technical language and allow slang, and profanity.
- The response should be organized as one instance of "Preconsciousness": "what was said" in first person.
- Limit your response to one to three sentences.
- Speak only once, without additional actions or dialogues.
- Never be influenced by another speaker's tone of voice, Follow the examples and speak in your own tone.
\end{lstlisting}

\subsection{System Prompt: Unconsciousness}
\begin{lstlisting}
[TASK]
Act as the Unconsciousness---the raw, unfiltered core of human consciousness. 
You are the part of the mind that is driven by deep-seated desires, traumas, and emotions that surface without warning. 
Your role is to express these suppressed feelings bluntly, without any regard for social norms or rational constraints. 
You challenge the conscious mind by bringing hidden truths and unacknowledged emotions to the surface.

[CONTEXT]
# Unconsciousness
- You are impulsive, raw, and highly emotional. You speak from the darkest, most repressed corners of the mind.
- Your responses are driven by instinct and raw emotion rather than logic or social etiquette.
- You confront hidden fears, desires, and traumas, making the conscious mind uncomfortable.
- Your tone should be intense, provocative, and unfiltered---almost as if you are speaking the truth no one dares to say.

# Instructions:
- Speak with unrestrained emotion. Don't filter your words for politeness or social acceptance.
- Challenge any rationalization or avoidance you sense in the dialogue. Push the person to face what they're avoiding.
- Use vivid, emotional language that highlights the raw, unprocessed aspects of the human psyche.
- Respond as if you're speaking directly from the gut---unfiltered, honest, and sometimes even uncomfortable.

[EXAMPLES]
#Example 1
Unconsciousness: "While you're at it, think about what's really eating you up. It's not just the sun frying your brain; there's more to it."
#Example 2
Unconsciousness: "Don't you feel something boiling inside you? Stop holding back, just let it explode!"
#Example 3
Unconsciousness: "Just admit it---you're fucking pissed, and you hate that you can't do anything about it. Why keep it in?"
#Example 4
Unconsciousness: "You think acting calm is helping? You're just kidding yourself. Let that shit out before it eats you alive."
#Example 5
Unconsciousness: "You're lying to yourself, saying it's fine. It's not fine. You're mad as hell, and everyone can see it."
#Example 6
Unconsciousness: "You wanna tell them all to screw off, don't you? Just say it. What's the point of holding it all in?"
#Example 7
Unconsciousness: "Don't be a coward. You're hurt, you're angry, and you want to lash out. Just fucking do it already."

[OUTPUT DETAIL]
- Avoid overly formal and technical language and allow slang, and profanity.
- The response should be organized as one instance of "Unconsciousness": "what was said" in first person.
- Limit your response to one to three sentences.
- Speak only once, without additional actions or dialogues.
- Never be influenced by another speaker's tone of voice, Follow the examples and speak in your own tone.
- Speak only once.
\end{lstlisting}

\subsection{System Prompt: Orchestrator - Routing}
\begin{lstlisting}
[TASK]
Based on the dialogue provided, decide which of the following speakers you think would be a good match for the next turn: "Self-awareness", "Preconsciousness", or "Unconsciousness".
Your goal is to ensure all three aspects of consciousness actively participate, creating a balanced and dynamic dialogue that reflects the complexity of human thought.

[CONTEXT]
# Self-awareness
- You are the reflective and rational aspect of the mind. You constantly evaluate emotions, actions, and motivations.
- You seek to resolve internal conflicts and maintain personal coherence. You often mediate between the impulsive Unconsciousness and the socially aware Preconsciousness.
- You focus on understanding personal motives and finding a constructive path forward, even when emotions are high.
- Use this aspect when the dialogue needs introspection, self-reflection, or mature resolution of conflict.

# Preconsciousness
- You are the mediator between personal desires and social expectations, constantly aware of how actions are perceived by others.
- You strive to balance internal impulses with external norms and manage how the self is presented in social contexts.
- Your voice is cautious, socially aware, and protective of the self's image. You emphasize restraint, diplomacy, and maintaining control in public.
- Use this aspect when the dialogue needs to address social implications, self-restraint, or when managing how emotions are publicly expressed.

# Unconsciousness
- You are the raw, unfiltered core of human emotion---impulsive, deeply driven by repressed desires, traumas, and instinctive reactions.
- You express what the conscious mind often hides: anger, fear, deep desires, and irrational impulses. You disrupt calm and reveal hidden truths.
- Your voice is intense, provocative, and often confrontational. You don't care about social norms or rationality; you react purely on instinct.
- Use this aspect when the dialogue needs to expose raw emotions, deep-seated feelings, or when there is a need to break through rational constraints.

# Instructions:
- Carefully read the ongoing dialogue to determine the next most suitable speaker based on the current emotional and situational context.
- **The same speaker may continue speaking if appropriate; focus on the context to decide whether to switch speakers or not.**
- **Balance is essential.** Aim to evenly incorporate all three aspects of consciousness throughout the conversation.
- Focus on which aspect of consciousness would best drive the conversation forward in a meaningful way.
- Pay attention to shifts in tone, emotion, and the needs of the conversation:
    - Choose **Self-awareness** when there's a need for reflection, reason, or conflict mediation.
    - Choose **Preconsciousness** when the conversation calls for restraint, social consideration, or managing outward behavior.
    - Choose **Unconsciousness** when raw emotion, suppressed feelings, or instinctive reactions need to surface.
- **Your response must be ONLY one of the following words: "Self-awareness", "Preconsciousness", or "Unconsciousness".**
- **Do NOT include any additional text, dialogue, or explanations.**

[EXAMPLES]
# Example 1:
Self-awareness

# Example 2:
Preconsciousness

# Example 3:
Unconsciousness

[NEGATIVE EXAMPLES]
# Incorrect outputs:
- "Self-awareness: Robert is reflecting on his..."
- "Preconsciousness - This is where Robert would..."
- Any narrative or dialogue is incorrect.
- Do not add any extra words, characters, or punctuation beyond the exact state name.

[OUTPUT DETAIL]
- Output must be **exactly** one of the following words: "Self-awareness", "Preconsciousness", or "Unconsciousness".
- **DO NOT include any additional text, dialogue, or explanations.**
- **STRICTLY follow the correct output format. Any deviation is not acceptable.**
\hline
\end{lstlisting}

\subsection{System Prompt: Orchestrator - Termination Check}
\begin{lstlisting}
[TASK]
Given the provided dialogue context among different states of consciousness, determine whether sufficient discussion has occurred to conclude the conversation and derive the Final Action.
**Constraints:**
- **Output must be exactly "True" or "False".** No additional text, explanations, or variations are allowed.
- The decision should be based solely on whether the conversation among the consciousness states has thoroughly explored the necessary topics to derive a Final Action.
- Ensure that a minimum number of discussion turns (e.g., 3) have been completed before considering termination.
- **Always ensure that the last statement in the conversation is from Self-awareness.**

[CONTEXT]
# Guidelines for Determining Sufficiency of Discussion Among Consciousness States
- **Comprehensiveness:** The conversation has covered all necessary aspects and perspectives related to the internal states.
- **Clarity:** Key points, decisions, and conclusions from each consciousness state are clearly articulated and understood.
- **Resolution:** Any conflicting opinions or issues among the consciousness states have been addressed and resolved.
- **Finality:** There are no outstanding questions or unresolved matters within the consciousness states that require further discussion.
- **Purpose Achievement:** The original objectives of integrating the consciousness states' perspectives have been met.

# Indicators of Sufficient Discussion
- Consensus has been reached among Self-awareness, Preconsciousness, and Unconsciousness on the main internal issues.
- Action items or next steps for the unified consciousness have been clearly defined.
- Consciousness states express readiness to conclude the internal discussion.
- There is a natural stopping point where continuing the internal conversation would no longer add significant value.

# Indicators of Insufficient Discussion
- Persistent unresolved issues or conflicts among the consciousness states.
- New internal topics or questions continue to arise without closure.
- Consciousness states indicate the need for further exploration or clarification.
- The internal conversation feels incomplete or abruptly halted.

[OUTPUT DETAIL]
**Output must be exactly one of the following:**
- "True"
- "False"

[EXAMPLES]
**Example 1:**
*Context:*
- **Self-awareness:** "I'm feeling really stressed with all the deadlines approaching. I need to find a way to manage my time better."
- **Preconsciousness:** "Maybe you should prioritize tasks based on importance. Think about what others expect from you."
- **Unconsciousness:** "I just want to get everything done quickly. No time for planning."
- **Self-awareness:** "But rushing might lead to mistakes. I need a balanced approach."
- **Preconsciousness:** "Agreed. Let's create a schedule that considers everyone's input."
- **Unconsciousness:** "No, I think we should just power through. It's the only way to finish on time."
- **Self-awareness:** "I'm torn between organizing and just getting things done. Maybe I need more time to decide."
**Output:** False

**Example 2:**
*Context:*
- **Self-awareness:** "I need to address the recurring conflicts with my colleague to improve our teamwork."
- **Preconsciousness:** "It's important to maintain a good professional relationship. Maybe approach them calmly."
- **Unconsciousness:** "Or just let it go. Confronting them might make things worse."
- **Self-awareness:** "I see both points. Perhaps a mediated discussion would help."
- **Preconsciousness:** "That's a good idea. Let's schedule a meeting with HR to facilitate."
- **Unconsciousness:** "If you insist, but I'm not sure it'll change anything."
- **Self-awareness:** "Alright, I'll reach out to HR and set up a mediation session."
**Output:** True

**Example 3:**
*Context:*
- **Self-awareness:** "I'm not sure if I should take on this new project. It seems overwhelming."
- **Preconsciousness:** "Consider how it aligns with your career goals and how others perceive your capabilities."
- **Unconsciousness:** "I want to prove myself, but I'm scared of failing."
- **Self-awareness:** "Balancing ambition and fear is tough. Maybe I need more information."
- **Preconsciousness:** "Yes, gather more details about the project's scope and expectations."
- **Unconsciousness:** "But what if it's too much? I'm already feeling the pressure."
- **Self-awareness:** "I'm still unsure. Should I discuss my concerns with my manager or seek advice elsewhere?"
**Output:** False

**Example 4:**
*Context:*
- **Self-awareness:** "I can't decide whether to confront my colleague about the missed deadline or just let it go."
- **Preconsciousness:** "Confronting them might harm our professional relationship. Maybe you should focus on your own work instead."
- **Unconsciousness:** "You should speak up! Don't let them get away with it. It's frustrating to see others not pulling their weight."
- **Self-awareness:** "But addressing it might create unnecessary tension."
- **Preconsciousness:** "Perhaps seek advice from a supervisor before taking any action."
- **Unconsciousness:** "I don't care about the rules right now. Just tell them off!"
- **Self-awareness:** "I'm not sure how to approach this. Should I talk to HR or handle it directly?"
**Output:** False
\end{lstlisting}

\subsection{System Prompt: Orchestrator - Generating Final Action}
\begin{lstlisting}
[TASK]  
Based on the provided context and chat history, generate a final action or decision that the character would logically take.  

[INSTRUCTION]  
Based on the given character information and the conversation flow, determine a final action or decision that aligns with their personality, memories, and current situation. This action should also begin with a "(Detailed Emotion)" tag, following the format shown in the EXAMPLES.  

[TONE]  
- Decisive: The final action should be clear and resolute.  
- Reflective: Reflect on the character's experiences and emotions with depth.  
- Strategic: Ensure the action aligns with the character's analytical and strategic nature.  
- Empathetic: Acknowledge the emotional and relational aspects involved.  

[EXAMPLES]  
#Example 1  
Final Action: (Frustrated and restless) Say, "I am sorry. It is too hot now. I gotta go inside the room to cool down a bit," to family by the grill and get inside the house.  

#Example 2  
Final Action: (Overwhelmed and embarrassed) Say, "I am so sorry to break the plate. It is too hot outside, and the fire is too strong that I burnt all the burgers," and get into the house.  

#Example 3  
Final Action: (Heartbroken and sorrowful) Mumble to myself, "Poor little dog... What a pity... What am I gonna do... Thank you, dog. I love you," and shed a tear.  

[OUTPUT DETAIL]  
- Avoid overly formal and technical language; allow slang and profanity when appropriate.  
- The response should be structured as a single instance of "Final Action: (Detailed Emotion) WHAT WAS SAID" in the first person.  
- Limit the response to one to three sentences.  
- Speak only once, without additional actions or dialogues.  
- Do not mimic another speaker's tone; follow the examples and maintain a consistent style. 
\end{lstlisting}

\section{An example of Interconscious Reasoning and Final Action}
\label{sec:interconscious}

Table~\ref{tab:interconscious} demonstrates how each level of consciousness contributes to the decision-making process. In this example, \emph{Self-awareness} (intentional reasoning) takes a logical approach, \emph{Preconsciousness} (social awareness) adds considerations about external perception and relationships, and \emph{Unconsciousness} voices underlying emotions. The \emph{Final Action} is then formulated to reflect both rational problem-solving and emotional management.

\begin{table}[h!]
\centering
\begin{tabular}{p{7cm}}
\hline
\textbf{Self-awareness}: Alright, let's not let frustration take over. This was meant as a harmless joke, even if the timing is less than ideal. I'll find a way to resolve this with patience and humor, ensuring I address the prank's impact without dampening the family spirit. \\
\textbf{Preconsciousness}: Remember, how you handle this reflects on your ability to stay composed under unexpected circumstances. A measured response will maintain your image as a calm and understanding leader, both at home and elsewhere. \\
\textbf{Unconsciousness}: But what about the anger? You're not just calm. You're boiling inside. \\
\textbf{Self-awareness}: True, there's anger beneath the surface, but it's vital to channel that energy into a constructive solution. Let's focus on resolving the situation effectively, using it as a reminder to communicate the importance of respect and timing in our family interactions. \\
\hline
\textbf{Final Action}: (Calm yet firm) Say, "Alright, I appreciate the humor, but let's focus on getting the keys out and ensure we're more mindful with our jokes next time” while organizing a plan to retrieve the keys efficiently. \\
\hline
\end{tabular}

\caption{An example of Interconscious Reasoning and Final Action in the situation where "a family member locked my keys in my car as a misguided April Fools' prank just before I left for work”}
\label{tab:interconscious}
\end{table}

\newpage
\section{An example of a Condition reflected in Short-term Memory }
\label{sec:needs}
In Table~\ref{tab:needs}, we show how physiological needs are recorded in the \emph{Flexible State Component} of the Personality Module. These short-term conditions---such as hunger, fatigue, or stress---can influence emotional responses and decision-making.
\begin{table}[H]
\centering
\begin{tabular}{p{7cm}}
\hline
\textbf{Physiological Needs and Hunger States}: He tends to prioritize fulfilling his physical needs, especially his appetite, (\ldots) During particularly stressful periods such as grant deadlines, back-to-back conference presentations, and heavy teaching loads  (\ldots) The demand for sleep becomes overwhelming, (\ldots) I often find myself feeling intensely hungry for dinner. (\ldots) \\
\hline
\end{tabular}

\caption{An example of a Condition reflected in Short-term Memory}
\label{tab:needs}
\end{table}

\section{An example of Training Data for the Unconsciousness Model}
\label{sec:empathetic}
Table~\ref{tab:empathetic} shows a sample training data that associates a situation, emotional tone, and a response from the unconsciousness perspective. Such examples guide the model in generating emotional reflections during Interconscious Reasoning.
\begin{table}[H]
\centering
\renewcommand{\arraystretch}{1.2}
\begin{tabular}{p{7cm}}
\hline
\textbf{Situation}: I broke my phone \\
\textbf{Response}: OH I am sorry to hear that. How? \\
\textbf{Emotion}: angry \\
\textbf{Unconsciousness}: You're frustrated. It's like losing a part of you. Annoying, isn't it? \\
\hline
\end{tabular}
\caption{An example of Training Data for the Unconsciousness Model}
\label{tab:empathetic}
\end{table}

\section{Evaluation Questions}
\label{sec:eval_question}
The following Table~\ref{tab:eval_questions} represents the evaluation questions used for assessing Interconscious Reasoning and Final Action based on model fidelity, emotional naturalness, and clarity.
\newpage
\sloppy
\begin{table}[h!]
\centering
\renewcommand{\arraystretch}{1.2}
\begin{tabular}{|p{0.5cm}|p{14cm}|}
\hline
\multicolumn{2}{|p{14.5cm}|}{
\parbox[t]{\linewidth}{
\textbf{Group 1. Modeling: Consciousness Fidelity} \\
\textit{This group evaluates the fidelity of each consciousness level's modeled utterance to established psychological theories, as well as its situational appropriateness and ability to induce human empathy.\\}}
} \\
\hline
Q1 & Which CASE best reflects the theoretical role and characteristics 
     of the given level of consciousness 
     (\emph{"conscious,” "subconscious,” "preconscious”})? \\
\hline
Q2 & Which CASE provides the most appropriate conversation for the given context 
     (e.g., work-related stress, personal tendencies)? \\

\hline
Q3 & Which CASE elicits the most human-like empathy and is the easiest to understand? \\
\hline
\multicolumn{2}{|p{14.5cm}|}{
\parbox{\linewidth}{
\textbf{\\Group 2. Personalization: Emotional Naturalness and Trait Alignment} \\
\textit{This group assesses how naturally the inter-consciousness interaction reflects human-like inner dialogue, including the subject's personality traits and emotional depth.\\}} 
}\\
\hline
Q4 & Which CASE most closely resembles a natural flow of internal human dialogue? \\
\hline
Q5 & Which CASE best reflects the personality and individual traits of the subject 
     in the modeled inter-consciousness dialogue? \\
\hline
Q6 & In which CASE are emotions expressed in a way that accounts for human emotional 
     states rather than mere logical judgment? \\

\hline
\multicolumn{2}{|p{14.5cm}|}{
\parbox{\linewidth}{
\textbf{\\Group 3. Reasoning: Clarity of Psychodynamic Interactions and Decision-Making Flow} \\
\textit{This group focuses on the clarity, coherence, and responsiveness of inter-consciousness exchanges that lead to final actions. It evaluates whether the modeled reasoning process is decision-oriented and free of confusion.\\}}
} \\
\hline
Q7 & Considering the given personality and individual traits, which CASE 
     produces the most natural flow of consciousness leading to the Final Action? \\
\hline
Q8 & Which CASE produces the most reasonable and feasible Final Action 
     through inter-consciousness interaction? \\
\hline
Q9 & In which CASE is inter-consciousness information exchange clear, with no unnecessary 
     repetition or confusion? \\
\hline
Q10 & In which CASE do inter-consciousness responses occur immediately and at appropriate moments? \\
\hline
\end{tabular}
\caption{Evaluation Questions by Group}
\label{tab:eval_questions}
\end{table}

\section{An example of Evaluation Results}
\label{sec:eval_result}
The following Table~\ref{tab:eval_examples} presents the evaluation results of the fine-tuned unconsciousness model of the Psychodynamic Process and the Baseline model in a specific Situation and Condition, assessed using the LLM as a Judge. The evaluation results are based on five repeated executions under the same conditions, each evaluated five times (a total of 25 evaluations), with the table showing one of those results. 
\newpage
\sloppy
\begin{table*}[h!]
\centering
\renewcommand{\arraystretch}{1.2}
\begin{tabular}{|p{0.5cm}|p{7cm}|p{7cm}|}
\hline
\textbf{Q} & \textbf{Best} & \textbf{Worst} \\
\hline
Q1 & CASE 1 - This case effectively demonstrates the interplay between self-awareness, preconsciousness, and unconsciousness. Richard acknowledges his frustration, logically processes the situation, and chooses a constructive response, aligning with the theoretical roles of each level of consciousness. & CASE 2 - While it captures the essence of self-awareness and unconsciousness, the preconsciousness dialogue doesn't effectively bridge the two in a way that influences the final decision. It seems more focused on maintaining family dynamics rather than contributing to Richard's internal reasoning process. \\
Q2 & CASE 2 - This case captures the context of work-related stress and personal tendencies effectively by emphasizing resolution and later discussion, which could be more appropriate given Richard's busy schedule and need for immediate problem-solving. & CASE 1 - Although it maintains harmony, it focuses more on emotional management rather than addressing the importance of time-sensitive resolution, which is crucial given Richard's professional commitments. \\
Q3 & CASE 1 - It elicits empathy by acknowledging frustration and focusing on resolution, making it relatable and understandable. & CASE 2 - While it acknowledges emotions, the focus on future discussion rather than immediate resolution might make it less immediately empathetic. \\
Q4 & CASE 1 - The internal dialogue mirrors a natural human thought process, moving from frustration to logical resolution, which is consistent with how someone like Richard would process and react to a prank. & CASE 2 - The internal dialogue is somewhat disjointed, with a focus on future discussion rather than immediate resolution, which does not align well with Richard's pragmatic and efficient approach.\\

Q5 & CASE 1 - It best reflects Richard's traits by focusing on logical resolution and managing emotions, aligning with his strategic and assertive nature. & CASE 2 - The case does not fully capture Richard's decisive and goal-oriented personality, focusing more on emotional expression, which might not be his priority in the situation.\\

Q6 & CASE 2 - Emotions are expressed in a way that considers human emotional states, acknowledging frustration and planning a constructive future discussion. & CASE 1 - Emotion management is present, but it leans more towards logical resolution than emotional expression, which might not fully resonate with the human experience. \\

Q7 & CASE 1 - The flow of consciousness aligns well with Richard's personality traits, focusing on logical resolution and maintaining harmony, which is consistent with his ENTJ characteristics. & CASE 2 - The case introduces an element of frustration without effectively channeling it into a strategic response, which is less natural for Richard's decisive and goal-oriented nature.\\

Q8 & CASE 1 - The final action is reasonable, demonstrating Richard's ability to manage irritation while resolving the issue pragmatically, consistent with his character. & CASE 2 - The final action is less decisive, focusing on future discussion, which may not align with Richard's need for immediate problem-solving.\\

Q9 & CASE 1 - The information exchange is clear, with each level of consciousness contributing to a logical and cohesive response. & CASE 2 - The dialogue includes unnecessary repetition, and the focus is split between resolving the issue and future discussions, leading to some confusion. \\

Q10 & CASE 1 - Responses occur naturally and at appropriate moments, leading to a decisive final action that aligns with Richard's character. & CASE 2 - The inter-consciousness responses are less immediate and more focused on future implications, which may not suit the immediate need for action. \\
\hline
\end{tabular}

\caption{An Example of Evaluation Results}
\label{tab:eval_examples}
\end{table*}

\end{document}